\documentclass{INTERSPEECH2023}
\usepackage[nolist]{acronym}
\usepackage{algorithm2e}
\usepackage{hyperref}
\usepackage{verbatim}
\usepackage{subcaption}
\usepackage{amssymb}
\usepackage{multirow}
\usepackage{xcolor}
\usepackage{enumitem}
\usepackage{dcolumn}
\usepackage{threeparttable}
\usepackage{booktabs}
\usepackage{url,footnote}
\usepackage{breqn}
\usepackage{here}
\usepackage{colortbl}
\usepackage{bbding}
\usepackage{booktabs}

\interspeechcameraready

\title{InterFormer: Interactive Local and Global Features Fusion for Automatic Speech Recognition}
\name{
    \begin{tabular}{c}
       \large ~Zhi-Hao Lai$^{1, 2}$,~Tian-Hao Zhang$^{1, 2}$,~Qi Liu$^{1, 2}$,~Xinyuan Qian$^{1}$, ~Li-Fang Wei$^{1, 2}$, \\ ~Song-Lu Chen$^{1, 2}$, ~Feng Chen$^{2, 3}$,~Xu-Cheng Yin$^{1, 2, *}$\thanks{* Corresponding author.}
   \end{tabular}
   \vspace{-10pt}
}
\address{
    $^1$\normalsize School of Computer and Communication Engineering, University of Science and Technology Beijing, Beijing 100083, China\\
    $^2$ \normalsize USTB-EEasyTech Joint Lab of Artificial Intelligence, University of Science and Technology Beijing, Beijing 100083, China\\
    $^3$ \normalsize EEasy Technology Company Ltd., Zhuhai 519000, China
}
\email{\{zhihaolai, tianhaozhang\}@xs.ustb.edu.cn, \{xinyuanqian, songluchen\}@xs.ustb.edu.cn}

\begin{document}
\newacro{BFIM}[BFIM]{bidirectional feature interaction module}
\newacro{SFM}[SFM]{selective fusion module}
\newacro{ASR}[ASR]{automatic speech recognition}
\newacro{CNN}[CNN]{convolutional neural network}
\maketitle

\begin{abstract}
The local and global features are both essential for \ac{ASR}. Many recent methods have verified that simply combining local and global features can further promote \ac{ASR} performance.
However, these methods pay less attention to the interaction of local and global features, and their series architectures are rigid to reflect local and global relationships.
To address these issues, this paper proposes InterFormer for interactive local and global features fusion to learn a better representation for ASR. 
Specifically, we combine the convolution block with the transformer block in a parallel design. 
Besides, we propose a \ac{BFIM} and a \ac{SFM} to implement the interaction and fusion of local and global features, respectively.
Extensive experiments on public ASR datasets demonstrate the effectiveness of our proposed InterFormer and its superior performance over the baseline models.
\end{abstract}
\noindent\textbf{Index Terms}: automatic speech recognition, bidirectional interaction, parallel structure, selective feature fusion

\section{Introduction}
\label{sec:intro}
The local and global features are both essential for \ac{ASR}.
It is well known that \ac{CNN} can effectively exploit local features while Transformer-based models are highly effective for capturing long-range feature relationships that constitute global representations due to the self-attention mechanism.
Therefore, the \ac{CNN} and transformer models have been the building blocks of \ac{ASR} \cite{kriman2020quartznet, pratap2019wav2letter++, yeh2019transformer, hori2020transformer, wang2020transformer, liang2021transformer}.
Unfortunately, global features are not well captured by \ac{CNN} due to its limited receptive field \cite{han2020contextnet, majumdar2021citrinet}. 
Whereas the details of local features are ignored by Transformer models \cite{zhao2019muse, mao2021dual, Fan2021MaskAN}.

To alleviate the above limitations, recent works \cite{han2020contextnet,zhao2019muse,gulati2020conformer,kim2022squeezeformer,Radfar2022,li2022uniformer} have tried to combine local and global features.
ContextNet \cite{han2020contextnet} simply adds squeeze-and-excitation \cite{hu2018squeeze} module in each residual block to capture long-distance feature dependencies. 
Conformer \cite{gulati2020conformer} proposes to add a convolution module that captures local features on top of multi-head self-attention to augment the Transformer.
These methods have shown that models combining local and global features outperform the models using local and global features separately in end-to-end \ac{ASR}.
\begin{figure} [t]
    \vspace{-12pt}
    \centering
    \includegraphics[scale=0.394]{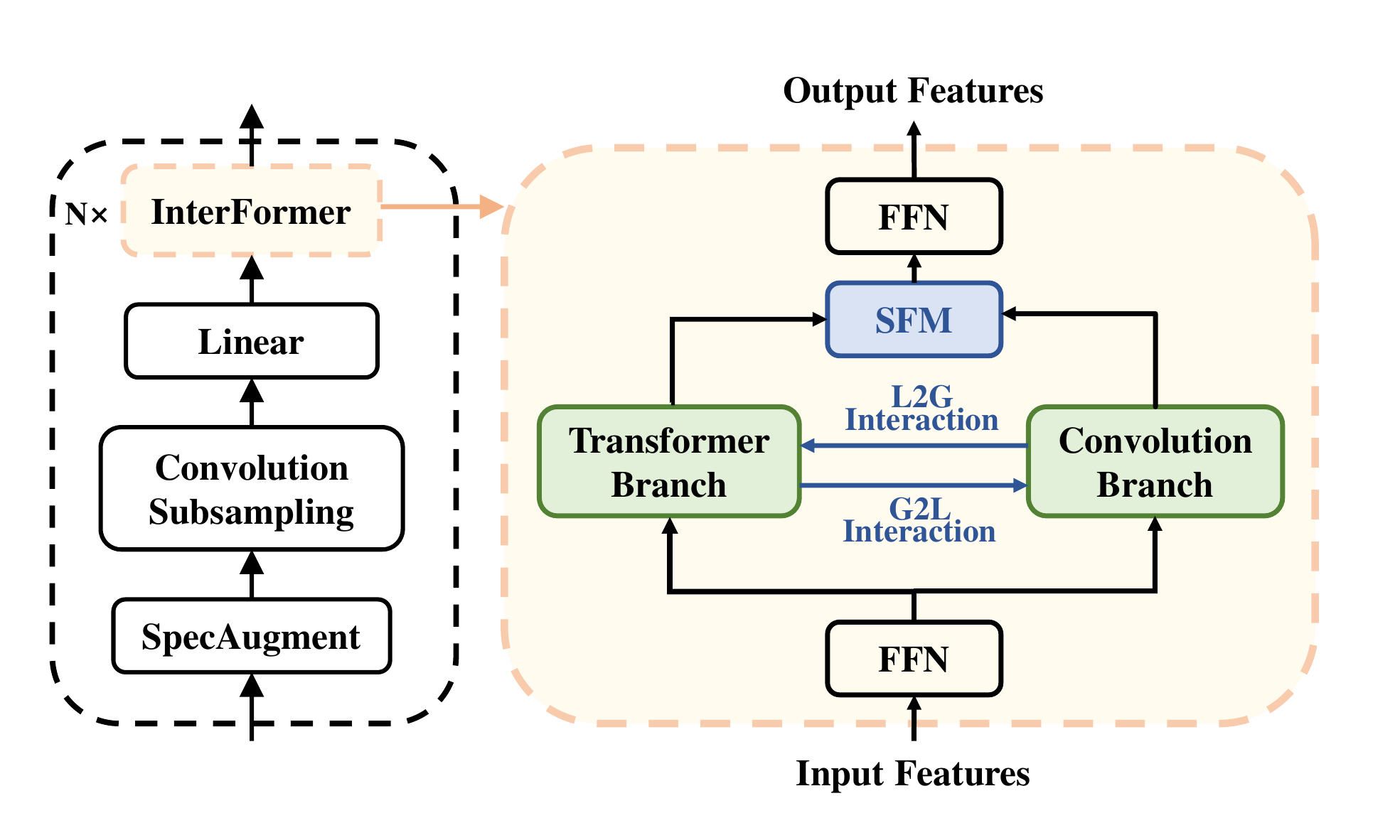}
    \caption{\textbf{The overall architecture of InterFormer.} It consists of two macaron-like feed-forward layers (FFN) and dual parallel branches. The blue arrows marked with L2G  Interaction and G2L Interaction.
    The resulting two-branch representations are then fused effectively in \ac{SFM}.
    N: number of identical InterFormer blocks
    }
\label{fig.overall}
\end{figure}

However, these structured serial methods rarely focus on the interaction of local and global features, which causes the following problems.
First, their serial architecture makes it hard to model various ranged dependencies and reflects local and global relationships in different layers, leading to latent information loss in the fusion process.
Second, the lack of effective interaction hinders further advances in feature representation ability. 

To address the aforementioned issues, in this paper,
we propose a parallel feature fusion network, namely InterFormer, 
to interactively fuse local and global features. 
As shown in Fig.~\ref{fig.overall}, our parallel structure consists of two branches. Convolution branch employs a convolution block to capture feature neighboring information, while Transformer Branch utilizes the multi-head self-attention (MHSA) with relative positional encoding to capture global dependencies.
Moreover, for the interaction between local and global features, \ac{BFIM} is designed as a bridge to provide complementary information for the transformer and the convolution branches, respectively, enhancing the feature representation learning ability in local and global aspects.
Taking into account the effective fusion of local and global features, \ac{SFM} is proposed to dynamically choose valuable characteristics based on the weight of the two branches. 
Such an interactive feature fusion method has a better representation capability to improve the \ac{ASR} performance. 

We conduct extensive experiments on three ASR datasets and compare the results with the state-of-the-art Transformer \cite{watanabe2018espnet} and Conformer \cite{watanabe2018espnet}.
Experimental results show that our InterFormer model achieves the best performance. Furthermore, extensive ablation studies demonstrate the effectiveness of our proposed \ac{BFIM} and \ac{SFM}.
\vspace{-0.1cm}
\begin{figure*} [htbp]
    \vspace{-12pt}
    \centering
    \includegraphics[scale=0.34]{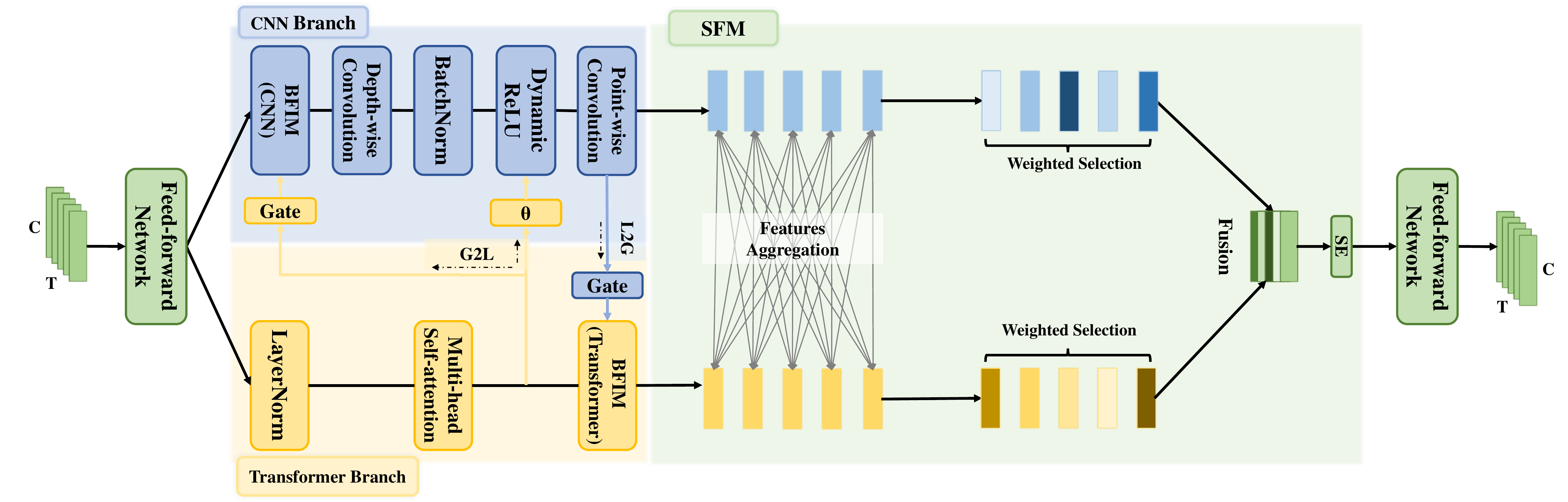}
    \caption{\textbf{The overall architecture of InterFormer.} Implementation details of the convolution branch, transformer branch, BFIM, and SFM. The convolution branch starts with a BFIM, followed by a single 1-D depthwise convolution layer, batchnorm, dynamic ReLU, and a pointwise convolution. The composition of the transformer branch is layernorm, a multi-head self-attention module and a BFIM. (SE: Squeeze-and-Excitation module; C: Channel dimension; T: Time dimension.)}
\label{fig.overall_detail_encode}
\end{figure*}

\section{InterFormer}
\label{sec:InterFormer with bidirectional interaction}
In this section, given an overview of InterFormer in Fig.~\ref{fig.overall}, we elaborate each internal block, as illustrated in Fig.~\ref{fig.overall_detail_encode}. 
Specifically, our model takes the raw audio waveform with a convolutional subsampling module to downsample the feature sequence, followed by the positional embedding addition. 
To take advantage of local and global representations for feature interaction, we employ a stack of N identical InterFormer blocks.

As shown in Fig.~\ref{fig.overall}, 
InterFormer replaces the serial connection {in Conformer} with a parallel connection. It consists of two branches where we extract local or global features using convolution operations or attention in each branch.
Considering the lack of interaction between two branches, we propose BFIM to interact local features in the convolution branch with global representations in the transformer branch.
To effectively fuse local and global information extracted from the two branches, SFM is proposed to weigh the importance of local and global features and dynamically select the most crucial characteristics.

\subsection{Parallel Branches for Extracting Local and Global Features}
Our InterFormer adopts dual parallel branches based on the standard self-attention and convolution elaborated as follows.

\textbf{Convolution branch.} we employ the convolution branch to extract local feature details.
As shown in Fig.~\ref{fig.overall_detail_encode}, the composition of the convolution branch is mainly similar to the convolution module of Conformer. It starts with a BFIM, which is a \ac{BFIM} used to implement the interaction of local and global information.
This is followed by a single 1-D depthwise convolution layer, batchnorm, dynamic ReLU, and a pointwise convolution.

\textbf{Transformer branch.} The transformer branch in InterFormer aims to capture the long-distance feature dependencies that constitute global representations. 
As shown in Fig.~\ref{fig.overall_detail_encode}, this branch consists of a MHSA module while adopting the relative positional encoding scheme \cite{dai2019transformer} and a BFIM block.

\subsection{Bidirectional Feature Interaction Module (BFIM)}
\label{ssec:BFIM}
Considering the lack of information interaction and the feature misalignment between convolution and transformer features, BFIM is designed as the bridge for information  exchange. 
The parallel branches contains the local-to-global (L2G) and global-to-local (G2L) interactions, respectively.
As shown in Fig.~\ref{fig.overall_detail_encode}, the information in the convolution branch flows to the other branch through the L2G interaction, strengthening the global features modeling ability.
Meanwhile, the G2L Interaction enables global features to flow from the transformer branch to the convolution branch, thus enhancing the local representation ability. 
As a result, the proposed bidirectional interactions provide complementary clues for each other and eliminate the feature misalignment between convolution and transformer features.

\begin{figure} [htbp]
    \centering
    \includegraphics[scale=0.28]{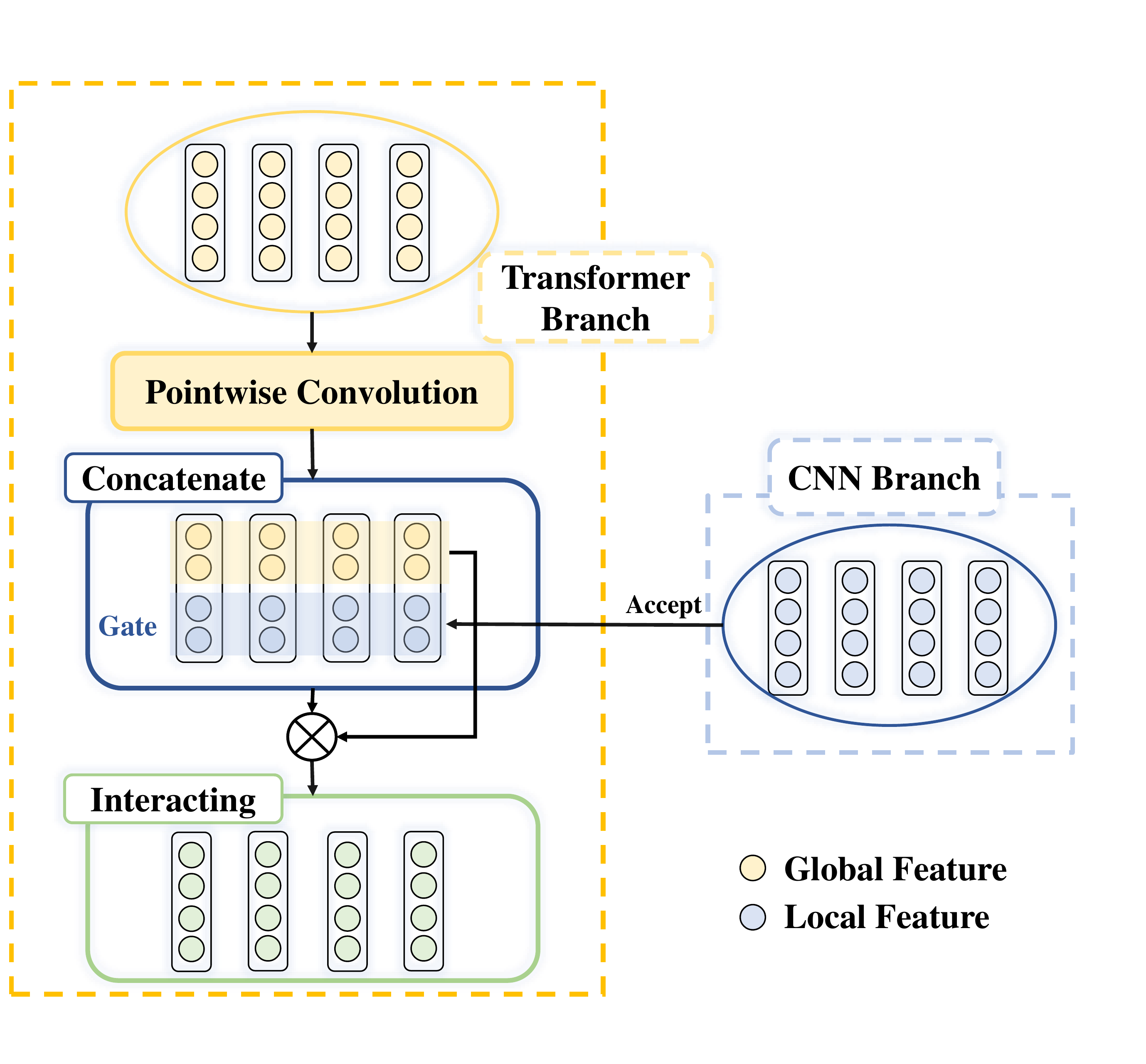}
    \caption{\textbf{Detailed design of the Interactions.} The Transformer branch accepts information from the CNN branch for interaction. Local features from CNN branch act as gating information to influence global features. The information flow is shown by the blue arrow in Fig.~\ref{fig.overall_detail_encode}.}
\label{fig.interact}
\end{figure}

\textbf{The L2G interaction.}
The input in the Transformer branch is passed through the attention mechanism to obtain the global feature $G$, followed by a layernorm and pointwise convolution. 
Then we concatenate the local feature $L$ and global feature $G$ where $L$ comes from the convolution branch, and the local information $L$ acts as gating information to influence global features.
The interaction process is depicted in Fig.~\ref{fig.interact}.
We compute the hidden features $h=\left \{  {{h}_{0},{h}_{1},...,{h}_{t}} \right \}$ in the convolution branch as:
\begin{equation}
\label{eq:interaction1}
h_{g}(x)=(P W \operatorname{Conv}(x)+b) \odot \sigma(L)
\end{equation}
where $P W \operatorname{Conv}$ is the pointwise convolution, $x \in R^{T \times d}$, 
$T$ is the sequence length, $d$ is the feature dimension, $b \in R^{T}$, $\sigma$ is the sigmoid function, and $\odot$ is the element-wise product between matrices. 

\textbf{The G2L interaction.}
For the G2L direction, we apply the same operation with the L2G direction to realize the interaction of local and global features:
\begin{equation}
    \label{eq:interaction2}
    h_{l}(x)=(P W \operatorname{Conv}(x)+b) \odot \sigma(G)
\end{equation}

To further exploit the global information for better information interaction, the following adjustments are applied to the activation function for the convolution branch.
Because the interaction between local and global information is a dynamic adjustment process, using the previous static activation function, such as swish and ReLU, may degrade the performance.
Therefore, we replace the previous static activation function with the dynamic ReLU activation function to achieve dynamic interaction of local and global features.
Specifically, the acquisition of $\theta$ is different from the original dynamic ReLU, which uses the original input $x$ to obtain. 
In our method, the output of $\theta$ is obtained through the global feature $G$ from the transformer branch. 
It applies two MLP layers (with a ReLU between them) and then uses a function to map the global information between -1 and 1. 
The process of dynamic ReLU is formulated as:
\begin{equation}
\label{eq:DyReLU}
\theta = 2\sigma(W_{2}\gamma\left(W_{1}G\right))-1
\end{equation}
where $\gamma$ is the ReLU activation, $W_{1}$ and $W_{2}$ denotes the MLP layer.
The above two methods achieve feature interaction. 

\vspace{-0.2cm}
\subsection{Fusion of Local and Global Features}
\label{ssec:SF Module}
Both local and global features are essential to improve the representation capability of the model.
To perform the fusion of local and global features, most previous methods use direct addition \cite{zhao2019muse, gulati2020conformer} or concatenation \cite{jiang2020convbert, peng2022branchformer}, which are too simple and effective
to selectively highlight the varying significance of the local and global features.
To weigh the importance of each branch in different layers and efficiently combine information from the dual branches, we propose \ac{SFM} to fuse local and global representations in a dynamic way.

\subsubsection{Direct Addition and Concatenation}
Let $Y_{conv}$ and $Y_{trans}$ be the output sequences of the convolution branch and transformer branch, respectively.
In the direct addition method, we add the features of the two branches along the time dimension directly:
\begin{equation}
    Y_{fused} = Y_{conv}+Y_{trans} 
\end{equation}
In the method of concatenation, we first concatenate the features of the two branches along the feature dimension and then use a linear transformation to make the result of the original dimension:
\begin{equation}
    Y_{fused} = Concat(Y_{conv},Y_{trans})W_{fuse} \in R^{2d \times d}
\end{equation}

\subsubsection{Selective Fusion Module (SFM)}

In the SFM, it weighs the importance of local and global features and selects the corresponding features based on the obtained attention weights.
From a macro perspective, the information obtained from the transformer branch is equivalent to a global receptive field acting on the transformer branch, while the information obtained from the convolution branch corresponds to acting on the convolution branch in a local receptive field.
Inspired by Selective Kernel Networks \cite{li2019selective}, we selectively fuse the two interacted features $L$ and $G$ from the convolution branch and transformer branch, respectively.
The core idea is to control whether the information of the interactions is retained or forsaken, as shown in Fig.~\ref{fig.overall_detail_encode}. To this end, we first concatenate features from two branches:
\begin{equation}
    X=Concat(L,G)
\end{equation}

Then, we calculate the mean value of each row of the input tensor in time dimension, followed by an MLP layer with a ReLU activation to reduce the channel dimension for better efficiency:
\begin{equation}
    \begin{array}{l}
        I=F_{m}=\frac{1}{T} \sum_{i=1}^{T} x_{i} \\
        \hat{X}=F_{f}=\gamma\left(W_{f}(I)\right)
    \end{array}
\end{equation}
where $\gamma$ is the ReLU activation, $W_{f} \in R^{c \times d}$  denotes the MLP layer.
Further, as shown in Eq.(\ref{eq:selective_fusion}), we use two MLP layers to act on $\hat{X}$ respectively to obtain two features that are restored to the original dimension and then concatenate them for information aggregation. 
These operations can generate a feature $Z$, guiding accurate and adaptive selections.
\begin{equation}
    \label{eq:selective_fusion}
    \begin{array}{c}
        Z=Concat\left(X_{1}=W_{u 1}(\hat{X}), X_{2}=W_{u 2}(\hat{X})\right)
    \end{array}
\end{equation}
where $W_{u 1} \in R^{d \times c}$, $W_{u 2} \in R^{d \times c}$.
It is well-known that the softmax function can introduce competition between attention weights, which is significant for obtaining selective attention. 
Thus, a softmax function is used to learn a selective factor $\alpha$, and it can select appropriate features from local and global aggregated features.
The output fusing convolution and transformer branch $X_f$ is generated by Eq.(\ref{eq:selective_fusion_res}):
\begin{equation}
    \label{eq:selective_fusion_res}
    X_{f}=Softmax(Z) \odot X
\end{equation}
Finally, to exploit contextual information, we use the Squeeze-and-Excitation module \cite{hu2018squeeze} to weigh the sum of the output by channel attention mechanism.

\section{Experiments}
\label{sec:Experiments}
\subsection{Experimental Setting}
\label{ssec:Experimental setting}
In this paper, we train and evaluate our proposed models on three public ASR datasets: Aishell-1 \cite{bu2017aishell}, Tedlium3 \cite{hernandez2018ted} and Librispeech \cite{panayotov2015librispeech}. Aishell-1 corpus has 170 hours of labeled speech collected from 400 speakers. 
Tedlium3 corpus was made from audio talks and contains 456 hours of 16kHz English TED talks. 
Librispeech consists of about 960 hours of 16kHz English read audiobooks. We choose them for experiments since they are well-known benchmarks for ASR tasks.

All our experiments are conducted on ESPNet toolkit \cite{watanabe2018espnet}, which uses PyTorch \cite{paszke2019pytorch} as a deep learning engine. For feature extraction, we use kaldi to extract input spectrograms of 80-dimensional mel-scale log filter banks computed from a 25ms window with a stride of 10ms. 
During acoustic front-end processing, we used SpecAugment \cite{park2019specaugment} and Speed Perturbation \cite{ko2015audio} on Aishell-1, for Tedlium3 and LibriSpeech dataset, we only use SpecAugment for data augmentation. The 2D-CNN front-end is comprised of 2 convolution layers with kernel size 3x3, filter 256, and stride 2 on the time dimension for 4x down-sampling. 
For different datasets, we adjust the attention head, attention dimension, hidden size and the number of encoder and decode layers to get better results. The Adam \cite{kingma2014adam} optimizer and warm up with  $\beta_{1}=0.9$, $\beta_{2}=0.98$, and $\epsilon=10^{-9}$. we employ label smoothing \cite{szegedy2016rethinking} of value 0.1 and dropout \cite{srivastava2014dropout} with probability 0.1 to prevent overfitting. Our experiments are conducted using 2 Nvidia 2080ti GPUs and use CER and WER as evaluation metrics, with the Chinese dataset using CER as an evaluation metric and the English dataset using WER as an evaluation metric.

\begin{table}
    \centering
    \caption{Experimental results present the Character Error Rate (in \%) of our proposed model and other previous models on the Aishell-1 Mandarin ASR task.}
    \label{tab:aishell}
    \arrayrulecolor{black}
    \scalebox{0.85}{
    \begin{tabular}{lccc} 
    \arrayrulecolor{black}\toprule
    Method & Dev & Test & Params(M)                \\ 
    \arrayrulecolor{black}\midrule
    ~ Citrinet\cite{majumdar2021citrinet}                    & 5.2                        & 5.71                        & 142                       \\
    \textit{ESPNet}\cite{watanabe2018espnet}                     &                            &                             &                           \\
    ~ Transformer                 & 6.0                        & 6.7                         & 29.7                      \\
    ~ Conformer                   & 4.6                        & 5.1                         & 46.2                      \\ 
    \midrule
    ~ InterFormer(ours)              & \textbf{4.4}                       & \textbf{4.9}                         & 46.8                     \\
    \bottomrule
    \end{tabular}
    }
\end{table}

\begin{table}[t]
    \caption{Word Error Rates (in \%) comparison among ASR models with Transformer, Conformer and InterFormer encoders. }
    \label{tab:English}
    \centering
    \scalebox{0.77}{
    \begin{tabular}{lccccc}
    \toprule
    Datasets & Evaluation Sets & Transformer & Conformer & InterFormer \\
    \midrule\midrule
    LibriSpeech & test/test-other & 4.2/10.5 & 4.1/9.2 & \textbf{3.7}/\textbf{9.0} \\
    \midrule
    Tedlium3 & test & 11.6 & 10.5 & \textbf{9.5} \\
    \bottomrule
    \end{tabular}}
\end{table}

\subsection{Main Results and Performance Analysis}
\label{ssec:Main Results and Performance Analysis}
In this section, we compare the proposed InterFormer model with other state-of-the-art models. For fairness, we reproduced Transformer and Conformer baselines with the same configurations, training, and decoding strategies.
All models are trained using the ESPNet toolkit without using any language model.

Table \ref{tab:aishell} compares the parameter sizes and character error rates (CER) on Aishell-1 dataset. The experimental results show that our model is superior to the Conformer and Transformer models.
Compared with the Conformer model, our model outperforms the Conformer by 0.2\%.

Table \ref{tab:English} compares WERs on LibriSpeech and Tedlium3. For the LibriSpeech dataset, because we use a configuration file with fewer parameters, without speed perturbation and language model, and fewer training epochs, the reproduced baselines are worse than those reported by other open-source toolkits.
When evaluated on test/test-other of the Librispeech dataset under the same configuration, the InterFormer outperforms Transformer and Conformer models with promising gains: +0.5/1.5\% and +0.4/0.2\%, respectively. 
When evaluated on the Tedlium3 dataset, the InterFormer achieve results of 9.5 on test outperforming the Transformer and Conformer.

\begin{table}
    \centering
    \caption{Ablation study of \ac{BFIM} on the Aishell-1 corpus. The baseline model in this table is Conformer with a successive design and has no interactions. $*$ means to remove dynamic ReLU.}
    \label{tab:Bidirectional interactions}
    \scalebox{0.85}{
    \begin{tabular}{c|cc|cc} 
    \toprule
             & \multicolumn{2}{c|}{Interaction} & \multicolumn{2}{c}{Aishell1}  \\
    Parallel & L2G & G2L                        & dev & test                    \\
    \midrule
             &     &                            & 4.6 & 5.1                     \\
    \Checkmark        &     &                            & 4.6 & 5.0                     \\
    \Checkmark        & \Checkmark   &                            & 4.5 & 5.0                       \\
    \Checkmark        &     & \Checkmark                           & 4.5 & 4.9                       \\
    \Checkmark        &     & \Checkmark$*$                          & 4.5 & 5.0                    \\
    \Checkmark        & \Checkmark   & \Checkmark                          & \textbf{4.4} & \textbf{4.9}                     \\ 
    \midrule
             & \multicolumn{2}{c|}{$\Delta$}           & \textbf{+0.2} & \textbf{+0.2}                     \\
    \bottomrule
    \end{tabular}
    }
\end{table}

\subsection{Ablation Studies}
\label{ssec:Ablation Studies}
In this section, we perform several detailed ablation studies to evaluate the contribution of each component and the effectiveness of different methods.
All models in ablation experiments are trained for 50 epochs on 2 Nvidia 2080 Ti GPUs with batch of 32 per GPU.

\subsubsection{Effect of BFIM }
\label{sssec:Effect of Bidirectional Feature Interaction Module}
Table \ref{tab:Bidirectional interactions} shows the results of the proposed \ac{BFIM}.
Starting from a Conformer block, We parallelize its convolution module and Transformer module and then continue to add various interactive components. According to the results, we can see that both L2G and G2L interactions outperform the model with only partial interaction or no interaction. 
We only use a simple and light-weight design for \ac{BFIM}, the gains are extraordinary, which shows that \ac{BFIM} can effectively provide important complementary clues for the convolution branch and the Transformer branch.

\subsubsection{Comparison of Different Fusion Operations}
\label{sssec:Comparison of different fusion operations}

Our proposed \ac{SFM} method and the other two fusion methods are discussed in Section \ref{ssec:SF Module}. Compared to the other two fusion methods, our proposed SFM has the ability to dynamically fuse the features of the convolutional and Transformer branches.
Table \ref{tab:diff_fusion} compares their performance on Aishell-1. According to the table information, we can see that our proposed \ac{SFM} is superior to the above two methods, which proves the effectiveness of our proposed fusion operations.

\begin{table}[t]
    \caption{Comparison of different fusion methods.}
    \label{tab:diff_fusion}
    \centering
    \scalebox{0.85}{
    \begin{tabular}{ccccc}
    \toprule
    \multicolumn{1}{c}{\multirow{3}{*}{Method}} & {\multirow{3}{*}{Interaction}} & {\multirow{3}{*}{Params(M)}} & \multicolumn{2}{c}{CER(\%)} \\ \cmidrule(r){4-5}
    \multicolumn{1}{c}{} & & & Dev & Test \\
    \midrule
    \multirow{1}{*} {Conformer}
    & \XSolidBrush &46.2   & 4.6 & 5.1 \\
    \midrule
    \multirow{1}{*} {Direct addition}
    & \XSolidBrush  &46.2  &4.7     &5.1      \\
    \multirow{1}{*} {Direct addition}
    & \Checkmark &46.5  &4.5     &5.0      \\
    \multirow{1}{*} {Concatenation}
    & \XSolidBrush  &47.8  &4.6     &5.1      \\
    \multirow{1}{*} {Concatenation}
    & \Checkmark  &48.4 &4.6     &5.0      \\
    \multirow{1}{*} {SFM(InterFormer)}
    & \XSolidBrush &46.6  &4.6     &5.0      \\
    \multirow{1}{*} {SFM(InterFormer)}
    & \Checkmark &46.8  &4.4     &4.9      \\
    \bottomrule
    \end{tabular}
    }
\end{table}

\section{Conclusion}
\label{sec:Conclusion}

In this paper, we propose InterFormer as an efficient dual parallel branches encoder for \ac{ASR}. 
Within InterFormer, we leverage the convolution branch to extract local features and the transformer branch to capture global representations. 
\ac{BFIM} promotes the feature interaction between the convolution branch and the transformer branch and alleviates the feature mismatch between the two branches. 
Moreover, we design a \ac{SFM} to fuse local features and global representations effectively, enhancing the modeling ability of audio representations. 
Experiments show that the proposed InterFormer outperforms the Conformer with comparable parameters. 
Furthermore, extensive ablation experiments have proved the effectiveness of \ac{BFIM} and \ac{SFM}.

\textbf{Acknowledgements: }The research is supported by National Key Research and Development Program of China (2020AAA0109700), National Natural Science Foundation of China (62076024, 62006018, U22B2055).

\bibliographystyle{IEEEtran}
\bibliography{mybib}

\end{document}